# Maximizing Information in Neuron Populations for Neuromorphic Spike Encoding

**Ahmad El Ferdaoussi, Eric Plourde, Jean Rouat**

NECOTIS, Department of Electrical and Computer Engineering, Université de Sherbrooke, Sherbrooke, QC, J1K 2R1 Canada

E-mail: `Ahmad.El.Ferdaoussi [at] usherbrooke [dot] ca`

July 22, 2024

**Abstract.** Neuromorphic applications emulate the processing performed by the brain by using spikes as inputs instead of time-varying analog stimuli. Therefore, these time-varying stimuli have to be encoded into spikes, which can induce important information loss. To alleviate this loss, some studies use population coding strategies to encode more information using a population of neurons rather than just one neuron. However, configuring the encoding parameters of such a population is an open research question. This work proposes an approach based on maximizing the mutual information between the signal and the spikes in the population of neurons. The proposed algorithm is inspired by the information-theoretic framework of Partial Information Decomposition. Two applications are presented: blood pressure pulse wave classification, and neural action potential waveform classification. In both tasks, the data is encoded into spikes and the encoding parameters of the neuron populations are tuned to maximize the encoded information using the proposed algorithm. The spikes are then classified and the performance is measured using classification accuracy as a metric. Two key results are reported. Firstly, adding neurons to the population leads to an increase in both mutual information and classification accuracy beyond what could be accounted for by each neuron separately, showing the usefulness of population coding strategies. Secondly, the classification accuracy obtained with the tuned parameters is near-optimal and it closely follows the mutual information as more neurons are added to the population. Furthermore, the proposed approach significantly outperforms random parameter selection, showing the usefulness of the proposed approach. These results are reproduced in both applications.

## 1. Introduction

Sensory systems use strategies of population coding in encoding stimuli. Populations of neurons with different properties thus collaborate to encode the perceptual features of the external stimuli. For example, sound is encoded by populations of auditory nerve fibers having low, medium, and high spontaneous rates [13]. Similarly, the coding of tactile pressure is done by populations of mechanoreceptive afferents that can be categorized into four types [4]. These population coding strategies improve the amount of relevant information from the stimuli that is encoded into spikes [27]. Extensive work in computational neuroscience has studied the correlations and functional interactions in population codes and how they can improve the coding of information [24, 3, 30, 15].

In spike-based processing systems, time-varying stimuli have to be encoded into spikes. There is a radical difference between the original stimuli and their representations with spikes: the stimuli signals are rich with information contained in their real-valued amplitude whereas spike trains are binary. Therefore, important information loss can occur in spike encoding, which can cause a bottleneck in the performance of neuromorphic applications [33]. To remedy this problem, population coding strategies are used in some works to encode more information on a stimulus signal than would be possible with a single neuron [18, 22].

In this context, population coding consists in encoding a stimulus signal into multiple spike trains rather than just one. Each spike train expands the available coding capacity and more information can thus be captured. It is straightforward to optimize the parameters of a population of neurons to maximize a performance metric in a given task (e.g. classification accuracy). A multitude of optimization algorithms, such as gradient descent, can be used to this end, but the optimality of the parameters found this way is application-dependent. By contrast, optimizing the encoding parameters using information theory has the advantage of being independent of the applications in which the spike-encoded data is used [9]. Moreover, it can also demand less computational resources than optimizing for a specific application if such an application involves a significant computational overhead to compute performance metrics such as classification accuracy [9]. To our knowledge, there are no works in the literature that seek to optimize the parameters of population codes using information theory for neuromorphic applications. As mentioned, the parameters are rather specifically optimized for a given task.

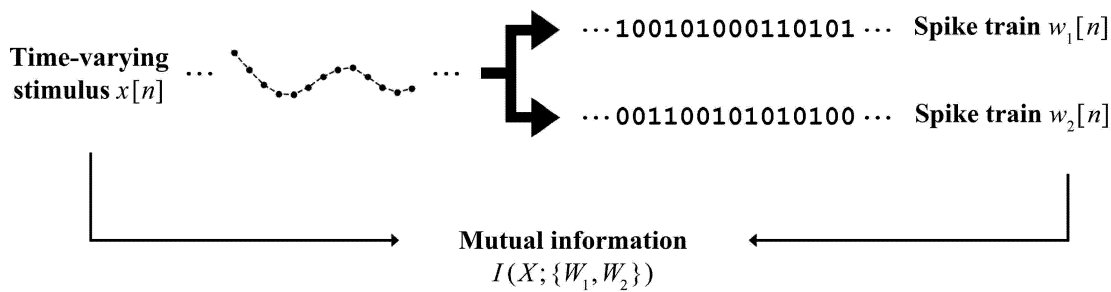

Figure 1: A time-varying stimulus signal $x[n]$ is encoded by two neurons into spikes trains $w_1[n]$ and $w_2[n]$. Random variables $X$, $W_1$, and $W_2$ characterize relevant aspects in the signal $x[n]$ and each of the spike trains $w_1[n]$ and $w_2[n]$. The mutual information that the two neurons encode jointly on the stimulus is $I(X; \{W_1, W_2\})$.

This work proposes an information-theoretic algorithm to tune the spike encoding parameters of a neuron population by maximizing the total information encoded from the stimuli [10, 9]. As shown in this work, by encoding a time-varying stimulus signal with a population of neurons, each neuron can capture unique information about the stimulus that other neurons in the population cannot. Aside from the unique contribution of each neuron in the population, information about the stimulus can further be encoded synergistically (i.e. only when the population is considered as a whole). By encoding more information from the input stimulus in this way, application performance can be improved.

The proposed algorithm is inspired by the Partial Information Decomposition (PID) framework [35, 20, 12] that decomposes multivariate information into partial "atoms" that make up the total information. The information encoded in the population is decomposed into the three partial "atoms": redundant information (present in more than one neuron), unique information (contributed uniquely by a single neuron), and synergistic information (that emerges only when the population of neurons is taken as a whole). The algorithm proposed in our work builds a population of neurons by iteratively finding the encoding parameters that maximize the information gained when adding a neuron to the population.

Two classification applications are presented. The first uses the PWDB database [6, 7] and consists in classifying simulated blood pressure pulse waves from 13 different body measurement sites. The second uses the SYNTH Monotrode database [26] and consists in classifying neural action potential waveform recordings simulated from 20 single units.‡

In both applications, the signals are encoded into spikes using a population a neurons. The encoding parameters are tuned using the proposed algorithm to maximize the mutual information between the signal and the spikes. Each additional neuron in the population increases the encoded information in the population. This, in turn, is reflected in gains in classification accuracy. These gains arise from the joint population code, beyond what could be accounted for by each neuron on its own in this context. Furthermore, the classification accuracy with the tuned parameters is found to be near-maximal in the sense that it is similar to the classification accuracy obtained when optimizing the parameters specifically to maximize classification accuracy in the task. The proposed approach also significantly outperforms configurations with random parameter selections.

‡ To avoid any confusion in this text, the words "neuron" and "spike" are reserved for neurons that are used to encode the data into spikes. The words "single unit" and "action potential" are used to refer to the units (neurons in the brain) from which the data (extracellular recordings of action potentials) of the SYNTH Monotrode dataset is simulated.

In the rest of this article, the proposed approach is presented, including the information-maximization algorithm for parameter tuning and the methodology of estimating the mutual information. Then, the two applications are presented along with the relationship between mutual information and classification accuracy. This is followed by a discussion and a conclusion.

## 2. Methods

This section presents the information maximization algorithm for parameter tuning and the methodology for estimating the mutual information between the signal and the spikes, as well as the example Leaky Integrate-and-Fire population code used in this work.

### 2.1. Proposed algorithm

Let $x[n]$ be a time-varying stimulus signal to be encoded into spikes with a population of neurons. Let $m$ be the total number of neurons in the population. The resulting spike trains are denoted $w_1[n]$, $w_2[n]$, ..., $w_m[n]$. A spike train in this context can be interpreted as a time-sampled signal that takes two values: 1 when there is a spike, and 0 otherwise.

*2.1.1. Case: $m = 1$ neuron* The encoding parameters of the first neuron are chosen to maximize the mutual information between the signal $x[n]$ and the first spike train $w_1[n]$, as characterized by the random variables $X$ and $W_1$, respectively. This mutual information is $I(X; W_1)$, and it represents the mutual dependence between the two variables, measured in units of bits [8]. The definition of the mutual information and the details of its estimation are presented in Section 2.2.

Multiple optimization strategies can be used to find the encoding parameters that maximize $I(X; W_1)$. The proposed algorithm is not limited to any one of them and it is not the objective of this work to compare their performance. Therefore, we use a grid search procedure for simplicity to study the potential of the approach.

Once the parameters for the first neuron are found, they are fixed. If the algorithm stops here, then the signal is encoded into a single spike train that maximizes information.

*2.1.2. Case: $m = 2$ neurons* Consider adding a second neuron to the population. The information encoded by the second neuron on its own is $I(X; W_2)$, whereas the information that is jointly encoded by both neurons is $I(X; \{W_1, W_2\})$ (Fig. 1).

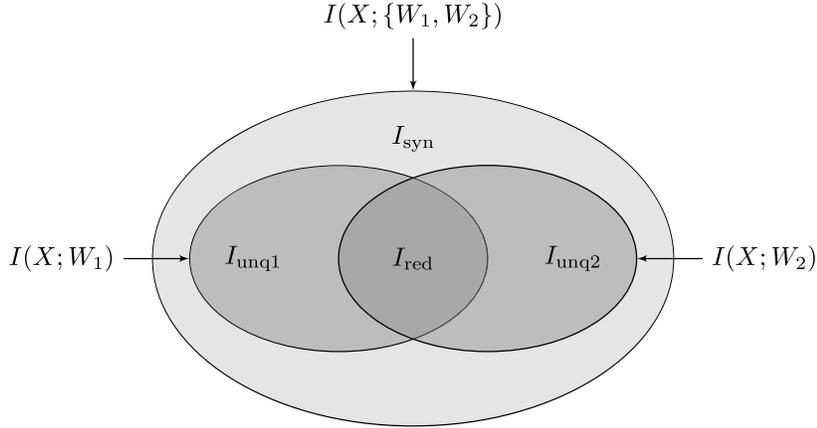

Figure 2: Partial Information Decomposition in the case of a population of two neurons encoding a stimulus signal. The total information encoded on $X$ by $W_1$ and $W_2$ is decomposed into three partial atoms: redundant, unique, and synergistic information. The redundant information $I_{\text{red}}$ is the information that overlaps between the two neurons, and it is contained in both $I(X; W_1)$ and $I(X; W_2)$. Each neuron contains unique information ($I_{\text{unq1}}$ and $I_{\text{unq2}}$) that is not present in the other neuron. When the two neurons are considered together, synergistic information $I_{\text{syn}}$ emerges that is not available when the two neurons are considered separately. (Figure inspired from [35].)

This joint information $I(X; \{W_1, W_2\})$ is composed of multiple constituent "atoms" in the Partial Information Decomposition framework (Fig. 2) [35]. The partial atoms that make up the total joint information are:

- Redundant information ($I_{\text{red}}$) that both neurons encode in copy.

- Unique information ($I_{\text{unq1}}$ and $I_{\text{unq2}}$) that each neuron encodes on its own.

- Synergistic information ($I_{\text{syn}}$) that is captured only when the two neurons are taken jointly. In other words, $I_{\text{syn}}$ is available in neither of the neurons separately, but it is available when they are considered together.

The relationships between the partial atoms of information are as follows:

$$I_{\text{unq1}} = I(X; W_1) - I_{\text{red}} \tag{1}$$

$$I_{\text{unq2}} = I(X; W_2) - I_{\text{red}} \tag{2}$$

$$I_{\text{syn}} = I(X; \{W_1, W_2\}) - I_{\text{unq1}} - I_{\text{unq2}} - I_{\text{red}} \tag{3}$$

The algorithm finds the encoding parameters of the second neuron by maximizing the joint information $I(X; \{W_1, W_2\})$. Since $W_1$ is fixed in this case and only $W_2$ is variable, this effectively maximizes the information contributed uniquely by the second neuron, as well as the synergistic information that emerges in the population.

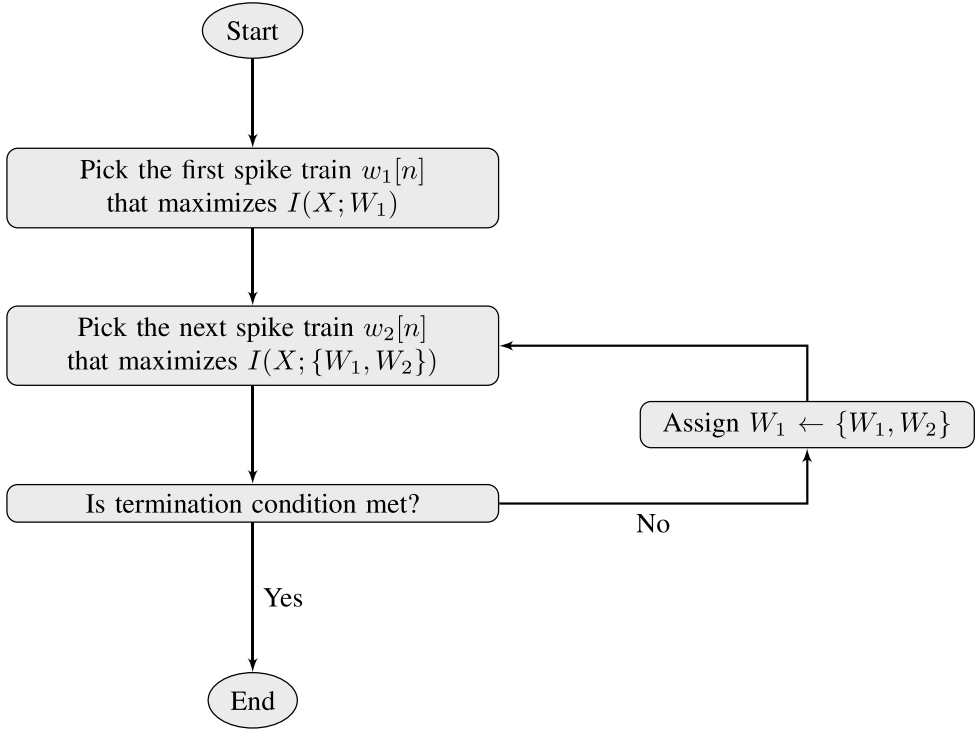

Figure 3: Flowchart of the algorithm proposed in this work to tune the parameters of a population of neurons encoding a time-varying stimulus signal. In every iteration, a new neuron is added to the population and its encoding parameters are chosen as the ones that maximize the total gained information.

*2.1.3. Case:* $m > 2$ *neurons* To add a third neuron to the population, the two previous neurons are considered as a fixed whole, and the algorithm is thus brought back recursively to the case of two neurons.

The general recursive algorithm is:

(i) Choose the encoding parameters of the first neuron as those that maximize the information $I(X; W_1)$ encoded in the first spike train $w_1[n]$.

(ii) Choose the encoding parameters of the new neuron to add as those that maximize the joint information $I(X; \{W_1, W_2\})$ encoded in the population as a whole (where $W_2$ is the random variable of the new neuron, and $W_1$ is the random variable of the previous neurons that are fixed in parameters).

(iii) Consider the two random variables $\{W_1, W_2\}$ as the new single random variable $W_1$ (i.e. assign $W_1 \leftarrow \{W_1, W_2\}$).

(iv) Repeat steps 2 and 3 to add new neurons to the population until a termination condition is met.

The algorithm is illustrated in Fig. 3. Several termination conditions are possible (e.g. a predetermined number of neurons in the population, a condition on the information gains, or a condition on the performance gains).

## 2.2. Mutual information estimation

In this work, the random variable $X$ is taken as the scalar amplitude of the time-varying stimulus signal $x[n]$. Since the amplitude of the signal is in general real-valued and not discrete, the random variable $X$ is a continuous variable, not a discrete one.

The amplitude of the signal is encoded in the timing of spikes in the spike trains. Let $w[n]$ be such a spike train and $W^T$ the random variable that represents these patterns of spikes in an observation window of duration $T$ in $w[n]$. $W^T$ is a discrete random variable as the spike train is binary.

The mutual information, measured in bits, is in this case a mixture between a continuous variable ($X$) and a discrete variable ($W^T$) [29]:

$$I(X; W^T) = \sum_w \int \log_2 \left( \frac{p(x, w)}{p(x)P(w)} \right) \mathrm{d}x \tag{4}$$

where probability density functions are denoted with a small $p(\cdot)$ and the probability mass function is denoted with capital $P(\cdot)$. The $k$-nearest neighbors estimator in [29] is used with the recommended $k = 3$ to estimate the mutual information.

Spike encoding might result in a latency being introduced such that, at a certain instant in time, the spike train variable $W^T$ is not maximally informative on the signal variable $X$. This latency might arise due to the algorithmic operation of the spike encoding technique on the time-varying signal. Due to this possible latency, the maximum information can actually be at an instant in the past. To take this latency into account, the spike train is successively shifted in the past by one sample at a time and the mutual information is estimated with each shift to find the maximum mutual information [21].

Information about an instantaneous signal amplitude value is contained in spike timing patterns in the spike train. The spike patterns closest in time to the amplitude value are the most informative. Spike patterns that are more distant contain less relevant information. Therefore, by using longer observation windows, more information can be captured on the instantaneous amplitude of the signal, in a diminishing returns fashion. This, however, is limited by the fact that windows of long duration lead to less reliable estimation of the mutual information because of the bias introduced by the finite sampling of data [14, 23].

To resolve the dependence of the mutual information on the duration $T$ of the observation window, a quadratic extrapolation is followed. Let $I(X; W)$ denote the "true" mutual information that the entire spike train encodes on the amplitude of the signal. Assuming that the spike train is long enough, this means that $I(X; W)$ is $I(X; W^T)$ in the limit when the duration of the observation window is very long (i.e. $T \to \infty$).

To extrapolate this value, the mutual information $I(X; W^T)$ is expanded to the second order of the inverse of $T$ [32, 34] with coefficients $a$, $b$, and $c$:

$$I(X; W^T) = a + \frac{b}{T} + \frac{c}{T^2} \tag{5}$$

$I(X; W^T)$ is then estimated for several durations $T$ of the observation window and the data is fitted to find the coefficients of the quadratic equation $a$, $b$, and $c$. The "true" mutual information $I(X; W)$ is then taken as the y-intercept in the quadratic extrapolation found with the function fit:

$$I(X; W) = \lim_{T \to \infty} I(X; W^T) = a \tag{6}$$

### 2.3. Spike encoding

Populations of Leaky Integrate-and-Fire (LIF) neurons [11] are used as an example in this work. Such populations of neurons with different firing thresholds have been used for example in the spike encoding of speech to achieve a rank-order coding scheme [18] and in silicon cochlea models that implement four LIF neurons with different thresholds to encode sound [17, 2]. Moreover, on the basis of information-theoretic efficiency metrics, LIF neurons were found to be more efficient [9] in encoding time-varying signals in comparison to other spike encoding techniques (like Send-on-Delta coding [19] and Ben's Spiker Algorithm [31]).

In this work, $x[n]$ is encoded into a spike train $w[n]$ with a discrete-time LIF neuron:

$$v[n + 1] = v[n] \times \delta + x[n] \tag{7}$$

where $v[n]$ is the membrane potential at discrete time index $n$, and $\delta$ is the decay constant. $\delta$ controls how much the membrane potential leaks at every time step, and it is comprised between 0 and 1. When $v[n]$ reaches a threshold $\theta$, a spike is generated and $v[n]$ is reset to zero (the resting potential). In this work, $\delta$ is taken to be 0.5.

The firing threshold $\theta$ modulates the number of spikes that are generated by the LIF neuron. If the threshold is sufficiently low, the neuron generates a spike at every time step and the spike train is therefore completely saturated with spikes. Conversely, if the threshold is sufficiently high, the neuron never generates spikes and the spike train is completely empty. For the purposes of parameter optimization, it is more useful to consider the action of the firing threshold $\theta$ on the resulting spike train than to investigate $\theta$ directly. The spike density $\rho$ captures this action of the firing threshold:

$$\rho = \frac{1}{N} \sum_{n=0}^{N-1} w[n] \tag{8}$$

where $N$ is the total number of time samples in the spike train. Spike density can be seen as a normalized spike count or a normalized firing rate, and it is comprised between 0 and 1.

In this work, the encoding parameter to optimize is thus chosen to be the firing threshold and it is investigated through the average spike density of the resulting spike trains.

## 3. Applications

This section presents two applications. The goal in both is to encode a dataset of stimuli with a population of neurons using the proposed approach and then classify the stimuli using the spikes in the neuron populations.

The first application consists in classifying simulated blood pressure pulse waves from different body measurement sites. The second application consists in classifying neural action potential waveforms from simulated single units from the brain.

In both applications, each stimulus consists of a single-channel time-varying signal. The stimuli are encoded into spike trains using populations of up to five neurons, with Leaky Integrate-and-Fire neurons used here as an example (see Section 2.3). The encoding parameters of the population are tuned using the proposed algorithm to maximize the information captured on the stimulus signals.

As mentioned previously, the goal is to classify the spike-encoded stimuli. Since the objective of this work is to study the relative gain in classification accuracy achieved by maximizing encoded information, a simple classifier with fixed parameters is used: Support Vector Machines (SVM) [5] with radial-basis functions. The features are extracted from the spike trains as follows. Each spike train is divided into 20 equally-spaced bins with 50% overlap. Each bin is then reduced to the average number of spikes contained within it. Therefore, classification is based on 20 features in the case of a single spike train, 40 features in the case of two spike trains, and so on. SVMs are implemented with the `scikit-learn` Python library [25]. The datasets are split into 80%/20% for training and test, respectively. No validation sets are considered since there are no classifier hyperparameters to tune.

In both tasks, the class of a stimulus can be decoded from the information present in the amplitude of its signal. More of this information can be captured in the spikes using a population coding strategy, and this can therefore potentially lead to higher classification accuracy.

After both applications are presented, the relationship between mutual information and application performance is discussed.

### 3.1. Application 1: Pulse wave classification

*3.1.1. Data* The Pulse Wave Database (PWDB) [6, 7] consists of simulated arterial pulse waves, representative of healthy adults, with a total of 4374 virtual subjects.

The database contains five types of features for every sample (arterial flow velocity, luminal area, flow rate, pressure, and photoplethysmogram pulse). The database is designed for "extensive *in silico* analyses of haemodynamics and the performance of pulse wave analysis algorithms" [6] and it is used for example in vascular age prediction for early cardiovascular risk detection [1].

In the application considered here, the task is to classify the blood pressure pulse waves in the 13 different measurement sites of the human body that they are simulated from. Fig. 4 shows three example pulse waves from 3 of the 13 different virtual measurement sites in the human body. The pulse waves are sampled at 500 Hz. The chance accuracy is 7.7%.

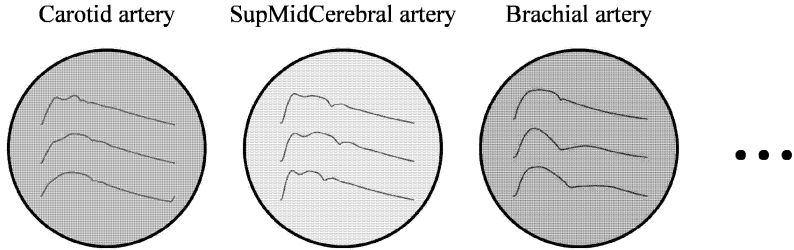

Figure 4: Application 1: Classification of blood pressure pulse waves simulated from 13 different measurement sites from the human body in the Pulse Wave Database (PWDB). The figure shows three example measurement sites (the carotid, the brachial, and the superior middle cerebral artery) with three example signals from each.

The task consists in encoding the blood pressure pulse waves into spike trains and classifying the spike count features extracted from the resulting spike trains. The LIF firing threshold is the encoding parameter to optimize using the information-maximization algorithm. A population of up to five LIF neurons is built with the algorithm. The mutual information is estimated from a representative subset of 260 randomly chosen pulse waves.

*3.1.2. Case: $m = 1$ neuron* The pulse wave stimuli are encoded by a single neuron into a single spike train each. A search procedure is run on the LIF firing threshold parameter to span the entire range of spike densities. The mutual information is estimated for different firing threshold values. Each firing threshold value yields a collection of spike trains that are characterized by their average spike density $\rho$.

The mutual information curve $I(X; W_1)$ as a function of the average spike density is shown in Fig. 5 (blue curve). $I(X; W_1)$ reaches a maximum of 3.058 bit (circled) for an average spike train density of 35%. The firing threshold for the first neuron in the population is thus fixed as the one that achieves this average spike density.

Classification in this case (single neuron in the population) achieves 45.1% accuracy.

*3.1.3. Case: $m = 2$ neurons* The firing threshold of the second LIF neuron in the population is found by maximizing the joint information $I(X; \{W_1, W_2\})$. Note that in this case, the first neuron has a fixed threshold, and it is the firing threshold of the second neuron that is varied. A search procedure is run on this firing threshold and the joint information is estimated for different threshold values.

The joint information curve $I(X; \{W_1, W_2\})$ as a function of the average spike density of the second LIF neuron is shown in Fig. 5 (red curve). This curve is shown atop the previous information curve $I(X; W_1)$. The two curves coincide at the average spike density of 35%, where the firing threshold of the second neuron becomes identical with the (fixed) firing threshold of the first neuron.

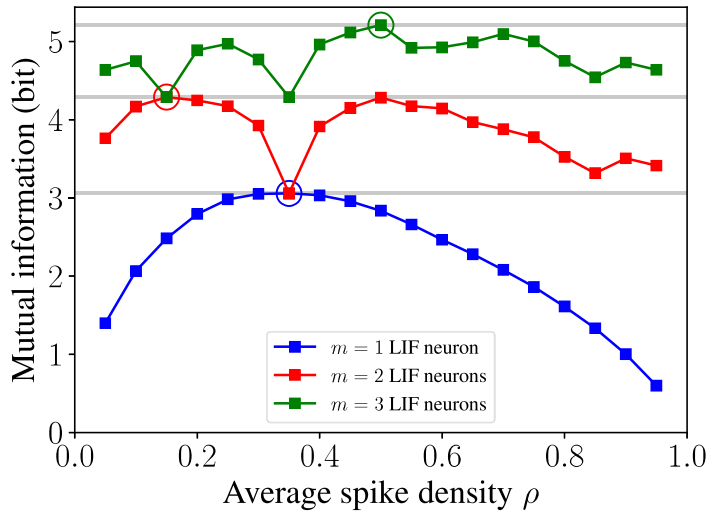

Figure 5: Application 1: Mutual information as a function of the average spike density. **Blue curve:** One neuron. A search is run on the LIF firing threshold to span the entire range of spike densities. The firing threshold is then fixed as the one that maximizes the mutual information (at an average spike density of 35%). **Red curve:** Two neurons. With the firing threshold of the first neuron fixed, another search is run on the threshold of the second neuron, and the joint information is maximized. The threshold for the second neuron is fixed as the one that maximizes that joint information at the average spike density (for the second neuron) of 15%. **Green curve:** The algorithm recursively finds the firing threshold for the third neuron in the population, which is the threshold that yields an average spike density of 50%.

The joint information $I(X; \{W_1, W_2\})$ reaches a maximum of 4.289 bit for an average spike density (for the second neuron) of 15%. The firing threshold for the second neuron is thus fixed as the one that achieves this average spike density. Adding this second neuron to the population leads to a 40% increase in the encoded information, which is a substantial amount of information that is contributed both uniquely by the second neuron and in synergy with the first neuron.

Classification in this case (of two neurons in the population) achieves an accuracy of 58.6%. This is an increase in accuracy of 13.5% compared to the case of a single neuron in the population. The gained accuracy reflects the increase in encoded information.

*3.1.4. Case: $m = 3$ neurons* To add a third neuron to the population, the recursive procedure of the algorithm is followed. The joint information curve is shown in Fig. 5 (green curve) as a function of the average spike density of the third neuron.

Table 1: Summary of the Pulse Wave Classification Task Results

| Number of neurons | $m = 1$ | $m = 2$ | $m = 3$ | $m = 4$ | $m = 5$ |
|---|---|---|---|---|---|
| Mutual information | 3.058 bit | 4.289 bit | 5.209 bit | 5.841 bit | 6.116 bit |
| Classification accuracy | 45.1% | 58.6% | 63.6% | 67.5% | 68.6% |

The joint information reaches a maximum of 5.209 bit for an average spike density of 50%. The firing threshold for the third neuron is thus fixed at the one that achieves this average spike density. Adding this third neuron to the population represents a gain in information of 21% compared to the case of two neurons, showing the diminishing returns on gained information (i.e. every neuron added to the population leads to a gain in information that is smaller than the gain from the previous neuron).

Classification in this case (three neurons in the population) achieves an accuracy of 63.6%, which is an increase of 5% compared to the case of two neurons in the population. The diminishing returns are thus also reflected in the performance.

*3.1.5. Summary* Table 1 recapitulates the mutual information and the classification accuracy results presented previously and reports additional results for the cases of four and five neurons. Adding neurons to the population leads to an increase in the encoded information which is in turn reflected in increased classification accuracy in the task.

*3.2. Application 2: Neural action potential waveform classification*

*3.2.1. Data* The SYNTH Monotrode database comprises numerous simulations lasting 10 minutes each of single-channel extracellular recordings of action potentials sampled at 24 kHz [26]. This database is frequently used in the benchmarking of spike sorting algorithms [16, 28].

In the application considered here, simulation number 10 is taken as a dataset. It includes simulated waveform recordings of action potentials from 20 single cell units. It also includes multi-unit recordings that combine two or more single units (Fig. 6).

The action potential waveforms (79 samples long) are labeled with the time of the action potential and the cell unit associated with it. All multi-unit action potential waveforms are assigned to the same class, which makes the total number of classes 21 (20 single units plus the multi-unit class). The classes are not balanced in this task: the number of waveforms in each class ranges from 118 (single unit 19) to 2187 (multi-unit class).

The task considered here consists in encoding these waveforms into spike trains and classifying the spike count features extracted from the spike trains. The chance accuracy is 4.8%. As in the previous application, the LIF firing threshold is chosen as the encoding parameter to optimize with the information-maximization algorithm. A population of up to five LIF neurons is built using the algorithm. The mutual information is estimated from a representative subset of 1000 randomly chosen action potential waveforms.

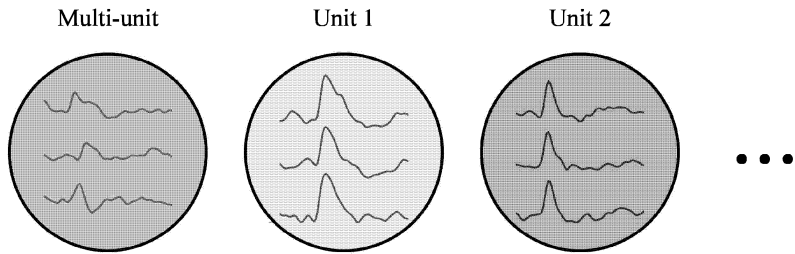

Figure 6: Application 2: Classification of action potential waveforms simulated from 20 single units from the SYNTH Monotrode dataset. The figure shows three classes (the multi-unit class, the class of single unit 1, and the class of single unit 2) with three example waveforms from each.

*3.2.2. Case: $m = 1$ neuron* The same methodology as in the previous application is followed. Fig. 7 shows the mutual information curves as a function of the average spike density of the spike trains.

In the case of a single neuron, $I(X; W_1)$ reaches a maximum of 1.876 bit for an average spike density of 15% (Fig. 7). The firing threshold of the first LIF neuron is thus taken as the one that achieves this average spike density. This configuration achieves a classification accuracy of 81.8%.

*3.2.3. Case: $m = 2$ neurons* A second neuron is added. $I(X; \{W_1, W_2\})$ reaches a maximum of 2.752 bit (47% increase in information) for an average spike density of 70% for the second neuron (Fig. 7). The firing threshold of the second neuron is thus taken as the one that achieves this average spike density. This configuration achieves a classification accuracy of 93.4%.

*3.2.4. Case: $m = 3$ neurons* Similarly, the mutual information in the case of three neurons in the population reaches a maximum of 2.982 bit for an average spike density of 40% for the third neuron (Fig. 7). This represents only an 8% increase in information, and this is reflected in the classification accuracy of 94%, which is only marginally better than in the case of two neurons.

*3.2.5. Summary* Table 2 recapitulates the mutual information and the classification accuracy results presented previously and reports additional results for the cases of four and five neurons. Adding neurons to the population leads again to an increase in the encoded information which is in turn reflected in increased classification accuracy in the task.

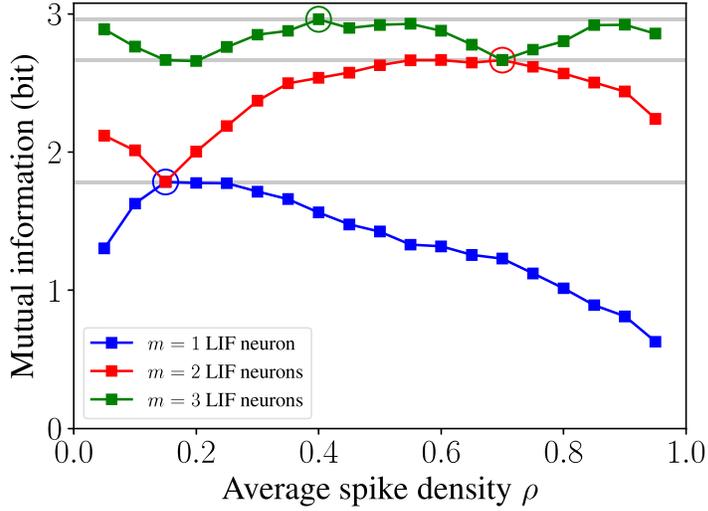

Figure 7: Application 2: Mutual information as a function of the average spike density. **Blue curve:** One neuron. The mutual information is maximized for an average spike density of 15%. **Red curve:** Two neurons. The joint information is maximized at an average spike density (for the second neuron) of 70%. **Green curve:** The joint information is maximized at an average spike density (for the third neuron) of 40%.

Table 2: Summary of the Neural Waveform Classification Task Results

| Number of neurons | $m = 1$ | $m = 2$ | $m = 3$ | $m = 4$ | $m = 5$ |
| --- | --- | --- | --- | --- | --- |
| Mutual information | 1.876 bit | 2.752 bit | 2.982 bit | 3.125 bit | 3.187 bit |
| Classification accuracy | 81.8% | 93.4% | 94.0% | 94.5% | 94.8% |

### 3.3. Relationship between mutual information and classification accuracy

In the following paragraphs, we study the interdependence between classification accuracy and maximization of the mutual information.

In the previous two applications, we performed an exhaustive search in the parameter space to find the encoding parameters that maximize the classification accuracy at every given number of neurons in the population. We compared their performance with the parameters found using the proposed information-maximization algorithm. Fig. 8 (bottom) shows this comparison (blue vs. green curves). It can be observed that the proposed algorithm reaches near-optimal accuracy.

It should be remembered that the parameters found using the information maximization approach did not involve any training or even any classification of the data in the applications. No models were considered either. The advantage here is to avoid the overhead when models are involved (e.g. the overhead of training spiking neural networks). The approach can thus be significantly faster and less demanding of computational resources.

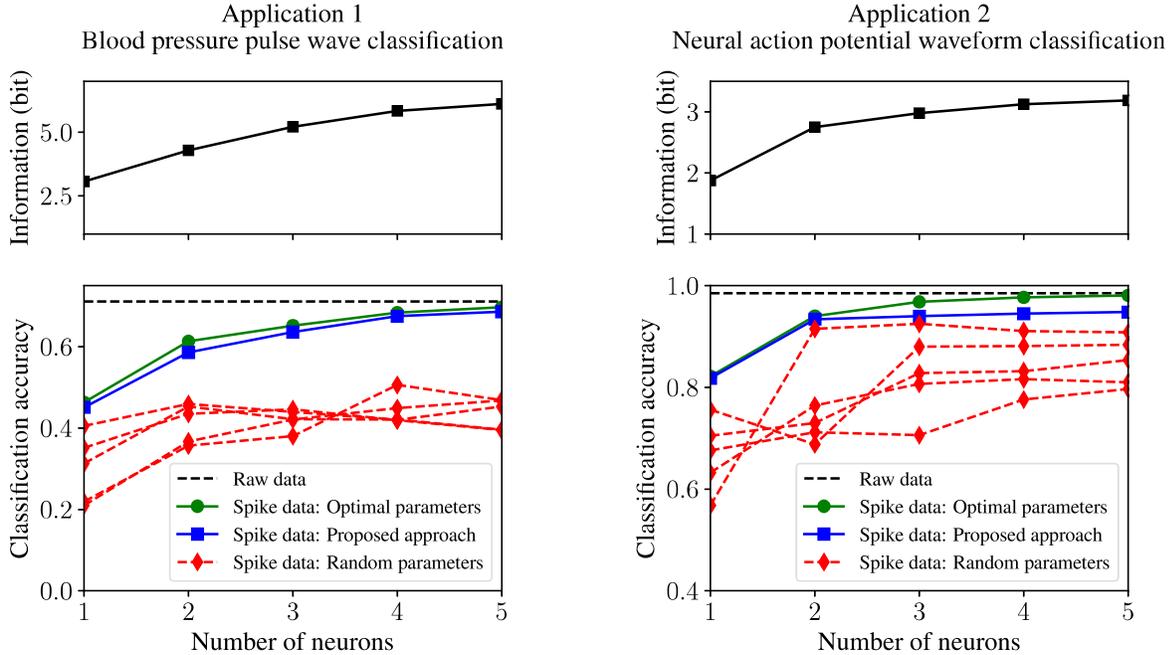

Figure 8: **Top:** Mutual information as a function of the number of neurons in the encoding population as determined by the algorithm proposed in this work for both applications. **Bottom:** Classification accuracy as a function of the number of neurons in the population. The curves are for the optimal parameters found through an exhaustive search (green curve), for the parameters found using the proposed information maximization approach (blue curve), and for configurations where the parameters are drawn randomly (red curves).

Furthermore, as a baseline of comparison, we evaluated the performance obtained with encoding parameters drawn randomly from the parameter space. We found that the proposed approach significantly outperforms random parameter selection. Fig. 8 (bottom) shows this comparison for five trials of random parameters (blue curve vs. red curves). From all the random trials, only one (in the second application) managed to achieve good performance through a lucky random choice of the encoding parameter of the second neuron. It can also be observed in several random trials that the performance can degrade in random parameter selection when adding more neurons to the population.

Additionally, the information gained by adding new neurons to the population closely follows the gains in performance observed in the applications. Fig. 8 (top) shows the evolution of the mutual information for both applications when adding new neurons to the population with the proposed algorithm. This can be compared with the evolution of the classification accuracy in the bottom figures (blue curves). Note that in this case the comparison between the mutual information and the classification accuracy should be made for each application separately, not between applications, as each application has its specificities.

## 4. Discussion

Both population codes and information theory are useful in spike-based processing systems. Populations of neurons can be used to capture more information on stimuli than would be possible with a single neuron. The gains in information can translate into significant gains in application performance. Information theory offers powerful tools to study the information content of neurons, independent of models and applications.

This work proposes an information-theoretic algorithm to tune the parameters of population codes that encode time-varying stimulus signals into spike trains. The algorithm builds a population of neurons iteratively by maximizing the total information encoded every time. The gains in encoded information translate to gains in task performance as seen in the two applications presented in this work. The performance is found to be near-maximal. Although a LIF population code is used in this work, the proposed algorithm is general and can be used with other population codes as well.

The advantage of encoding a single signal with a population of neurons is particularly relevant in cases where it is important to encode as much information as possible on the signal. This is the case for example in the applications presented in this work, where the data is single-channel. Note that in applications with multi-channel data, it might not be as useful to encode each channel into multiple spike trains. This is because the multi-channel nature of the data already constitutes a population coding strategy in itself. There also might be significant redundancy between channels. For these reasons, it might be sufficient to encode each channel with a single neuron in cases of multi-channel data.

In cases where this approach is relevant, mutual information can be used to find an initial parameter configuration that gives high performance. This configuration can then serve as a baseline for further tuning with the goal of improving accuracy, reducing spike density, or both. Using mutual information is also particularly relevant in tasks that are resource-heavy, where an exhaustive parameter search to optimize task performance itself is not tractable. Moreover, estimating the mutual information does not depend on the models used in specific applications.

The algorithm proposed in this work can also be used to gauge the number of neurons needed in an encoding population by using mutual information as a surrogate for task performance. It should be noted, however, that even though maximizing the mutual information means overall better encoding of the amplitude of the time-varying stimulus signals, this does not guarantee the best performance in the task [9]. Indeed, the performance in the classification tasks depends on a number of factors beyond the accurate encoding of amplitude (e.g. the classification model used, the choice of features to extract from the spike trains, the split of the data into training and test sets).

## 5. Conclusion

Although population codes have been used to encode stimuli for neuromorphic applications, they have never been specifically optimized to maximize the information encoded from the point of view of information theory. In fact, traditional approaches usually optimize the parameters of population codes for a specific application. Using information theory has the advantage of being independent of models and applications and having therefore the potential of being faster and less demanding of computational resources. The algorithm proposed in this work (inspired by the Partial Information Decomposition framework [35]) iteratively tunes the spike encoding parameters of a population of neurons by maximizing the total information encoded in the population. The maximization of information translates to gains in application performance as seen in the two applications presented in this work. The encoding parameters found through the proposed algorithm achieve near-maximal performance in the two classification applications and they significantly outperform random parameter selection. Thus, the algorithm proposed in this work can be useful in optimizing population spike encoding for neuromorphic applications and it can lead to more efficient designs for signal-to-spike encoding.

### Acknowledgments

This work was supported by the *Fonds de Recherche du Québec – Nature et Technologie* (FRQNT) and the Natural Sciences and Engineering Research Council of Canada (NSERC). This research was enabled in part by support provided by *Calcul Québec* (`calculquebec.ca`) and the Digital Research Alliance of Canada (`alliancecan.ca`). Thanks to Alexis Mélot for suggesting the neural neural action potential waveform classification task.